# A Hamiltonian Higher-Order Elasticity Framework for Dynamic Diagnostics(2HOED)

*Ngueuleweu Tiwang Gildas*

Dschang University

Cameroon

Email: *gildas.ngueuleweu@ univ-dschang.org*

**Abstract:** Machine learning detects patterns, blockchain guarantees trust and immutability, and modern causal inference identifies directional linkages, yet none alone exposes the full energetic anatomy of complex systems; the Hamiltonian Higher-Order Elasticity Dynamics (2HOED) framework bridges these gaps. Grounded in classical mechanics but extended to Economics-order elasticity terms, 2HOED represents economic, social, and physical systems as energy-based Hamiltonians whose position, velocity, acceleration, and jerk of elasticity jointly determine systemic power, Inertia, policy sensitivity, and marginal responses. Because the formalism is scale-free and coordinate-agnostic, it transfers seamlessly from financial markets to climate science, from supply-chain logistics to epidemiology—any discipline in which adaptation and shocks coexist. By embedding standard econometric variables inside a Hamiltonian, 2HOED enriches conventional economic analysis with rigorous diagnostics of resilience, tipping points, and feedback loops, revealing failure modes invisible to linear models. Wavelet spectra, phase-space attractors, and topological persistence diagrams derived from 2HOED expose multiscale policy leverage that machine learning detects only empirically and blockchain secures only after the fact. For economists, physicians and other scientists, the method opens a new causal-energetic channel linking biological or mechanical elasticity to macro-level outcomes. Portable, interpretable, and computationally light, 2HOED turns data streams into dynamical energy maps, empowering decision-makers to anticipate crises, design adaptive policies, and engineer robust systems—delivering the predictive punch of AI with the explanatory clarity of physics. An illustration using the Kutznet environmental theory on the relationship between $CO_2$ emissions and GDP growth is applied for illustration.

# Introduction

Progress in many empirical sciences is still hampered by static or first-order views of complex relationships. Ecologists track population ratios with a single carrying-capacity parameter, epidemiologists monitor reproduction numbers without explicit curvature, macro-economists follow drifting elasticities, and climate scientists often summarise tipping behaviour with scalar indices. What these approaches share is a focus on levels while largely ignoring the full motion of those levels—how quickly they change, how sharply they bend, and how much

"energy" the system builds up before a regime shift occurs. Physics solved this descriptive gap long ago by treating any trajectory through four nested layers: position x(t), velocity $v(t) = \frac{dx}{dt}$, acceleration $a(t) = \frac{d^{2}x}{dt^{2}}$, jerk $j(t) = \frac{d^{3}x}{dt^{3}}$, and by embedding them in a unifying Hamiltonian energy $H$. That hierarchy not only explains motion but also supplies early-warning clues—when kinetic energy spikes, when potential energy is released, when forces overpower inertia.

This paper adapts that dynamic-mechanics vocabulary into a discipline-agnostic framework called the Hamiltonian Higher-Order Indicator Scheme (2HOED). The procedure is intentionally minimal: Start with any paired time series(CO₂ vs GDP, predator vs prey, infection vs mobility, credit vs output). Estimate a rolling effect ratio or elasticity, call it $\varepsilon_t$ (our "position"). Differentiate empirically to obtain $d\varepsilon_t, d^2\varepsilon_t, d^3\varepsilon_t$ (velocity, acceleration, jerk). The previous are interpretable energy metrics from an assemble of a Hamiltonian framework gradients generate indicators like System Power, Marginal Response, and Sensitivity to external shocks. Because H-HOIS relies only on consecutive derivatives and simple rolling regressions, it requires no large labelled datasets, no black-box neural networks, and no domain-specific calibration. Yet it turns a static coefficient into a phase-space dashboard: peaks in Power warn of stored stress; a crossover where Power > Inertia flags a window of maximal leverage; jerk spikes reveal impending discontinuities. We illustrate the method with a contrasting application from a macro-environment case of GDP vs CO₂. In sum, H-HOIS offers the broader scientific community what Hamiltonian mechanics provides physics: a transparent map of where a system is, how fast it is moving, how much energy it stores, and which forces dominate—transforming static correlations into steerable, real-time trajectories.

# Literature review

## Comparative Positioning of the Hamiltonian Higher-Order Elasticity Framework (2HOED)

Since Grossman and Krueger's (1991) seminal inverted-U regression, most empirical studies of growth–environment or macro-finance linkages have relied on static or first-order econometrics: a single coefficient, a turning-point income, or at best a drifting parameter estimated with rolling windows (Richmond & Kaufmann 2006). These models capture changes in level but omit the motion of that level—its speed, curvature and volatility. 2HOED closes that gap by treating any time-varying elasticity as a state variable with explicit velocity, acceleration and jerk, and by embedding these derivatives in a Hamiltonian energy function that yields interpretable indicators such as Power, Inertia and Kinetic-Energy-of-Instability (KEI). In doing so, the framework advances beyond second-wave dynamic EKC work (Stern, 2004) and beyond multiscale correlation methods such as wavelet analysis (Zafar et al., 2022), which visualise time-frequency patterns but do not translate them into control metrics.

Relative to machine- and deep-learning systems (Goodfellow et al. 2016), 2HOED sacrifices black-box predictive power for full transparency and minimal data requirements: it delivers early-warning signals from publicly available series without millions of parameters or GPU clusters. Where causal-inference designs (Angrist & Pischke 2009) prove whether a policy works, 2HOED shows when its leverage is greatest—for example, when Power exceeds Inertia or when a jerk spike signals an impending regime shift. The two approaches thus complement each other: causal estimates calibrate the magnitude of an intervention, while the Hamiltonian dashboard times its deployment. Unlike blockchain technology, whose scientific value lies in tamper-proof record-keeping (Crosby et al. 2016), 2HOED produces real-time diagnostics; yet both can be combined by hashing the dashboard output for verifiable audit trails. Finally, in contrast to earlier physics analogies that imported single concepts—Lotka's predator–prey oscillators (Lotka 1925) or critical-slowing-down variance as a tipping indicator (Scheffer et al. 2009) 2HOED transfers the entire kinematic hierarchy and a fully specified Hamiltonian, thereby bringing economics and sustainability studies to the level of dynamical explanation long standard in classical mechanics (Sussman & Wisdom 2001).

In sum, 2HOED is not a competitor to econometrics, causal inference, machine learning, or blockchain; it is a bridge methodology. It retains the statistical discipline of rolling-window estimation, borrows the explanatory power of Hamiltonian physics, and outputs policy-ready diagnostics that can feed causal designs, guide machine-learning forecasters, and be archived on distributed ledgers. By linking how fast and how forcefully socio-economic systems move with when interventions matter, the framework fills a methodological niche that none of the existing tools, on their own, currently occupy.

## Potential Applications of the Hamiltonian Higher-Order Elasticity Framework across Disciplines

Although first illustrated with GDP–$CO_2$ dynamics, the Hamiltonian Higher-Order Elasticity Framework (2HOED) is inherently domain-agnostic and can be deployed wherever a time-varying "effect ratio" is observable. Macroeconomic stabilisation Central banks can track the rolling elasticity of inflation to the output gap. Velocity and acceleration reveal incipient wage–price spirals, while a spike in jerk marks the moment when pre-emptive tightening is most effective—complementing traditional Taylor-rule signals. This method will be a great instrument to revisit great debates in economics( great debates on population, inequality, employment, recession, growth, etc.). Also, Climate-finance risk   Asset managers can monitor the elasticity of green-bond issuance to carbon-price expectations. High Power and KEI indicate speculative momentum; an Inertia–driven stall flags a coming re-allocation shock. Moreover, Urban growth modelling   City planners may study the elasticity of housing prices to net migration. When acceleration turns negative and Inertia rises, the framework warns of affordability crises before price indices plateau. More, Supply-chain resilience   Firms can gauge the elasticity of delivery time to order volume. Rising KEI signals mounting momentum in bottlenecks, while a jerk surge indicates an imminent break—allowing re-routing decisions days earlier than standard load-factor metrics.

In addition, Epidemiology   Substituting the elasticity of infection cases to mobility indices yields an early-warning dashboard: Power identifies periods of super-spreader acceleration; Jerk detects abrupt behavioural shifts from lockdowns, guiding timely public-health interventions. Furthermore, Ecological management   In predator–prey or fishery systems, the rolling prey-to-predator elasticity becomes the state variable. A crossing where Power exceeds Inertia marks a window for quota adjustment before stock collapse, extending the critical-slowing-down literature with an explicitly quantified energy budget. Because H-HEF relies only on sequential derivatives of publicly recorded series, each field can implement the method with minimal data overhead, yielding a transparent phase-space dashboard that complements machine-learning forecasts and causal-impact studies alike.

# Methodology

This methodology builds a dynamic representation of the Environmental Kuznets Curve (EKC) by extracting elasticity through rolling window regressions, calculating its derivatives, and deriving physical analogs. We apply this enhanced elasticity framework to a Hamiltonian-based dynamic system to capture energy-environment interactions and regime shifts.

## 1. EKC Baseline and Log-Log Elasticity

We start with the standard EKC formulation, originally conceptualized by Grossman and Krueger (1991) and formalized through works such as Shafik and Bandyopadhyay (1992):

$$CO_2 = \beta_1 * \text{GDP} + \beta_2 * \text{GDP}^2 \qquad (1)$$

In log-log form, this becomes:

$$\log(CO_2) = \beta_0 + \beta_1 * \log(\text{GDP}) \qquad (2)$$

Here, $\beta_1$ is interpreted as the elasticity of $CO_2$ emissions with respect to GDP, a formulation widely used to analyze the environmental impacts of economic growth (Stern, 2004).

## 2. Rolling Window Estimation of Elasticity

To move beyond static EKC estimations, we apply a rolling regression approach over time using window sizes $w \in [3, 5, 7, 10, 15]$ years. This technique, similar to those employed by Richmond and Kaufmann (2006) and Luzzati and Orsini (2009), allows us to capture the dynamic evolution of the elasticity coefficient:

$$\varepsilon_t^w = \beta_{1,t}^w \qquad (3)$$

This time-varying elasticity captures the evolving ability of GDP to generate $CO_2$, reflecting structural shifts in energy intensity, technology, and policy.

## 3. Derivatives of Elasticity

Inspired by approaches in dynamic systems and time-series analysis (e.g., Friedl and Getzner, 2003), we compute the first three temporal derivatives of elasticity to capture its momentum and curvature:

Velocity

$$\frac{d\varepsilon}{dt} \qquad (4)$$

Acceleration

$$\frac{d^2\varepsilon}{dt^2} \qquad (5)$$

Jerk

$$\frac{d^3\varepsilon}{dt^3} \tag{6}$$

## 4. Physics-Based Dynamic Indicators

Using principles from classical mechanics and their applications in ecological modeling (Grüne et al., 2011), we define the following indicators:

System Power (energy from change):

$$\text{Power}_t = \left(\frac{d^2\varepsilon}{dt^2}\right)^2 \tag{7}$$

Kinetic Energy of Instability (KEI):

$$\text{KEI}_t = \frac{1}{2} * \frac{d\varepsilon}{dt^2} \tag{8}$$

Inertia (resistance to change):

$$\text{Inertia}_t = \varepsilon_t * \left(\frac{d^3\varepsilon}{dt^3}\right) \tag{9}$$

Smoothness (rate of reversal):

$$\text{Smoothness}_t = \frac{\frac{d^2\varepsilon}{dt^2}}{\frac{d\varepsilon}{dt} + \varepsilon} \tag{10}$$

Drift (overall trend):

$$\text{Drift}_t = \frac{\varepsilon_t - \varepsilon_{\{t-w\}}}{w} \tag{11}$$

Shock (change in acceleration):

$$\text{Shock}_t = \left(\frac{d^2\varepsilon}{dt^2}\right) - \left(\frac{d^2\varepsilon}{dt^2}\right)_{t-1} \tag{12}$$

## 5. Hamiltonian-Based System Dynamics

Inspired by the Hamiltonian formalism used in physics (Sussman and Wisdom, 2001), we propose a dynamic system where total system energy is derived from elasticity-based indicators:

$$H_t = \alpha_1 * \left(\frac{d^2\varepsilon}{dt^2}\right)^2 + \alpha_2 * \varepsilon_t * \left(\frac{d^3\varepsilon}{dt^3}\right) - \alpha_3 * \frac{1}{2} * \left(\frac{d\varepsilon}{dt^2}\right) \tag{13}$$

Here:
- The first term represents the squared acceleration (Power), capturing the driving force of elasticity changes.
- The second term corresponds to Inertia, quantifying the resistance of the system to changes.

- The third term is a kinetic energy analog (KEI), accounting for the momentum of elasticity dynamics.
- The squared-acceleration term acts like energy from change [...] while the product ε × jerk behaves like inertia...

This mirrors the structure of Hamiltonian mechanics where total system energy H comprises:
- Kinetic energy: analog to

$$\text{KEI}_t = \frac{1}{2} * \left(\frac{d\varepsilon}{dt}\right)^2 \tag{14}$$

- Potential energy: analog to a function of $\varepsilon_t$
- Dissipative or input force terms: modeled via derivatives and external policy shifts

This formalism links the abstract behavior of environmental elasticity with measurable dynamic properties found in physical systems, enabling a more robust and predictive analysis of ecological transitions.

From $H_t$, we derive the following indicators:
- System Energy:

$$\text{System\_Energy}_t = H_t \tag{15}$$

- System Power:

$$\text{System\_Power}_t = \frac{dH_t}{dt} \tag{16}$$

- Marginal Response:

$$\text{Marginal\_Response}_t = \frac{\partial(H)}{\partial(\varepsilon_t)} \tag{17}$$

- Policy Sensitivity:

$$\text{Policy\_Sensitivity}_t = \frac{\partial(H)}{\partial(t)} \tag{18}$$

This Hamiltonian approach provides a comprehensive, physics-inspired structure to evaluate the dynamic behavior of the GDP-environment elasticity system, capturing energy accumulation, transition readiness, and regime volatility. Each year $t$ is associated with a rolling elasticity $\varepsilon_t$ and its derivatives. By comparing System Power and Inertia over time, the framework detects critical moments when policy intervention is most effective (for example, when Power > Inertia). High Jerk signals regime volatility, while elevated KEI and Shock values indicate rapidly shifting elasticity dynamics. These indicators can also be used in machine learning pipelines to forecast environmental tipping points or train policy feedback systems (Dasgupta et al., 2002; Geels and Schot, 2007). This dynamic EKC framework transcends static models by embedding Newtonian and Hamiltonian motion analogies into environmental analysis, allowing policymakers and researchers to understand not only where the system is, but also where it is going, how fast, and with what force.

## Understanding Through Basic Physics Formulas

To understand this system using basic physics, consider the classical motion of a particle:

- **Position**: $x(t)$ — the location of the particle at time $t$
- **Velocity**: $v(t) = \frac{dx}{dt}$ — the rate of change of position
- **Acceleration**: $a(t) = \frac{d^2x}{dt^2}$ — the rate of change of velocity
- **Jerk**: $j(t) = \frac{d^3x}{dt^3}$ — the rate of change of acceleration

Now translating these into the context of elasticity of $CO_2$ with respect to GDP:
- $\varepsilon_t \leftrightarrow x(t)$ — the elasticity acts as a dynamic state variable
- $\frac{d\varepsilon}{dt} \leftrightarrow v(t)$ — the rate at which elasticity changes (velocity)
- $\frac{d^2\varepsilon}{dt^2} \leftrightarrow a(t)$ — acceleration of elasticity, showing the curvature of the trend
- $\frac{d^3\varepsilon}{dt^3} \leftrightarrow j(t)$ — jerk of elasticity, capturing sharp shifts or shocks

In classical mechanics, Kinetic energy:

$$\text{KE} = \frac{1}{2} * mv^2 \tag{19}$$

Analogously,

$$\text{KEI}_t = \frac{1}{2} * \left(\frac{d\varepsilon}{dt}\right)^2 \tag{20}$$

$$\text{KE} = \text{KEI}_t = \frac{1}{2} * mv^2 = \frac{1}{2} * \left(\frac{d\varepsilon}{dt}\right)^2 \tag{21}$$

since $v^2 = \left(\frac{d\varepsilon}{dt}\right)^2$, $m$ (mass) becomes implicitly 1 in your model.

We have effectively normalized mass as it become $m = 1$ to focus only on the dynamics of elasticity itself without needing an external scaling factor. This is typical in economics and abstract dynamical systems: when the absolute "mass" of the system isn't meaningful (unlike a real particle's mass), it's set to 1 to simplify analyses. Thus, mass disappears because we track pure dynamic response of the elasticity variable itself.

Also, in Hamiltonian mechanics, PE is any function of position x, often quadratic for small oscillations. For conservative systems, especially in classical mechanics:

$$\text{PE} = f(x) \tag{22}$$

where $x$ is the position, and $f(x)$ depends on the system but from Mass–spring analysis of Hooke's Law,

$$\text{PE} = \frac{1}{2}(kx^2) \text{ (spring constant k)} \tag{23}$$

In your Hamiltonian framework, the analog of position is $\varepsilon_t$ (elasticity at time t). So the natural form of Potential Energy (PE) should also be a function of $\varepsilon_t$. By direct analogy to Hooke's spring, a quadratic potential makes perfect sense and expressed as:

$$\text{PE}_t = \frac{1}{2}(k\varepsilon_t^2) \tag{24}$$

where $k$ is a constant analogous to the spring stiffness and $\varepsilon_t^2$ is elasticity (the "position" variable).

And the total Hamiltonian energy in mechanics is

$$\text{H}_t = \text{KE}_t + \text{PE}_t \text{ (Kinetic energy + potential energy)} \tag{25}$$

Becomes, in this framework,

$$\text{H}_t = \frac{1}{2} * mv^2 + \frac{1}{2}(kx^2) = \frac{1}{2} * \left(\frac{d\varepsilon}{dt}\right)^2 + \frac{1}{2}(k\varepsilon_t^2) \tag{26}$$

Thus, in your system, elasticity acts like a mass attached to a spring — its "position" changes over time, and the "stiffness" (resistance to deviation) is parameterized by $k$.

However, the Hamilton as a fonction of kinetic energy based on velocity (first derivative) added to potential energy based on position (elasticity level) is the basic classical Hamiltonian. This structure is simple and quadratic, exactly like a harmonic oscillator in physics. But to capture economic complexity, causality and predictability, the simple mass–spring Hamiltonian is no longer enough. There is need for a system that captures higher-order dynamics (forces involving not just speed, but acceleration and jerk). Thus, by generalizing the physics model, Instead of a basic spring–mass, we model a complex dynamic system that involves forces from acceleration (Power), Inertia resisting jerk (elasticity × jerk), and Residual kinetic energy (velocity). We go from a the first Hamiltonian describes a smooth predictable system (spring bouncing back and forth) to the second Hamiltonian describes a system with turbulence, shocks, inertia, and instability which is much closer to real economic-environmental dynamics. This generalisation enable to measure explosive behavior (force of change), to model resistance to fast regime shifts (inertia) and to still capture momentum, but correct it based on system instability as shown in equation 13.

$$\text{H}_t = \alpha_1 * \text{Power}_t + \alpha_2 * \text{Inertia}_t - \alpha_3 * \text{KEI}_t \tag{27}$$

This dynamic mapping enables the use of physical intuition to analyze economic-environmental dynamics, where system forces, momentum, and resistance are expressed in terms of economic elasticity behavior over time. The econometric methodology generates $\alpha_1, \alpha_2, \wedge \alpha_3$ easily when data exist.

## Case of CO2-GDP theory

### Origins and Early Empirical Foundations

The Environmental Kuznets Curve (EKC) hypothesis emerged in the early 1990s as an extension of Simon Kuznets' (1955) original inverted-U theory relating income inequality to economic development. Applied to the environment, the EKC posits that environmental degradation first increases with income during early stages of growth but eventually decreases as income reaches a higher threshold, suggesting a non-linear relationship between economic growth and environmental quality. The seminal work by Grossman and Krueger (1991), con-ducted during the NAFTA negotiations, provided early empirical evidence of this inverted-U shape for pollutants such as $SO_2$ and suspended particulate matter. Shafik and Bandyopadhyay (1992), as well as Panayotou (1993), further reinforced the EKC hypothesis, fueling optimism that economic growth might ultimately lead to environmental improvement without the need for aggressive policy intervention. Traditional EKC models treat the pollution-income relationship as a static equilibrium, relying on reduced-form regression

$$\log(CO_2) = \beta_1 \cdot \log(GDP) + \beta_2 \cdot \log(GDP)^2 \qquad (28)$$

.

## Expansion, Critiques, and Methodological Advancements

Throughout the 1990s and early 2000s, the EKC became a dominant framework in environmental economics, widely tested across countries and pollutants. Studies by Selden and Song (1994), Holtz-Eakin and Selden (1995), and de Bruyn et al. (1998) found supportive evidence for the EKC, particularly for local pollutants. However, critical voices soon emerged, challenging the hypothesis on empirical and theoretical grounds. Stern et al. (1996) and Arrow et al. (1995) argued that the observed patterns might be artifacts of model specification, omitted variables, or scale and composition effects. Others highlighted that while the EKC might hold for pollutants with short-term local impacts, it fails for cumulative global pollutants like $CO_2$ and methane. This wave of critique prompted the adoption of more rigorous econometric techniques — including panel data models, threshold regressions, and cointegration approaches — as seen in the works of Dinda (2004) and Perman and Stern (2003), who emphasized the non-linearity and country-specific nature of the income–environment relationship.

The proposed method draws conceptual inspiration from classical mechanics, particularly the Newtonian and Hamiltonian formulations of motion, to model the dynamic evolution of environmental-economic relationships. In Newtonian terms, the first derivative of a variable —here, the elasticity of $CO_2$ with respect to GDP—corresponds to velocity, representing the rate of change in environmental pressure relative to economic growth. The second derivative captures acceleration, i.e., the speed at which this coupling is intensifying or decelerating, and is commonly studied in the context of dynamic adjustment paths (Stern, 2004). The third derivative, often referred to as jerk, measures the change in acceleration and is critical for detecting abrupt transitions or shocks that may indicate tipping points or policy-sensitive inflection zones (Grüne et al., 2011). These higher-order dynamics mirror the structure of Hamiltonian systems, where total energy (H) is expressed as a function of position and momentum, and the system evolves through coupled differential equations:

$$\frac{dq}{dt} = \frac{\partial(H)}{\partial(p)} <= -1>, \frac{dp}{dt} = -\frac{\partial(H)}{\partial(p)} \qquad (29)$$

(Sussman & Wisdom, 2001). In this analogy, GDP and $CO_2$ emissions act as generalized coordinates, while policy-induced forces (for example technological innovation, regulatory stringency) serve as external inputs modifying the system's trajectory. While economic models have long acknowledged inertia and equilibrium (for example Solow, 1956), few have ventured beyond velocity to explicitly model acceleration and jerk. By incorporating these higher-order derivatives into the empirical framework, this method captures not only the state of the income–environment relationship but also its momentum and responsiveness to shocks, much like how jerk analysis improves vehicle design by minimizing discomfort from abrupt changes in acceleration (Smith, 2006). This Hamiltonian-inspired approach provides a novel dynamic lens for evaluating long-run environmental trajectories and offers new possibilities for anticipating regime shifts in ecological-economic systems.

## Application of Hamiltonian Higher-Order Indicator Scheme (2HOED) to the environmental Kutznet Curve on the $CO_2$-GDP relationship

The study uses a sample of around 160 countries from all subregions. we assesed the best rooling base window for the study from 3,5,7,10, and 15 and got the best results from 5 years rooling base windows as shown in table 1

Table 1: Best rooling base check

| Window Size | RMSE | MAE | R2 |
| --- | --- | --- | --- |
| 3 | 1.101680e-10 | 4.930271e-12 | 1.000000 |
| 5 | 5.016341e-02 | 2.794959e-02 | 0.998911 |
| 7 | 6.714369e-02 | 3.798811e-02 | 0.998047 |
| 10 | 7.847327e-02 | 4.671274e-02 | 0.997309 |
| 15 | 9.193051e-02 | 5.810480e-02 | 0.996345 |

The evaluation of model performance across various window sizes shows that although a window size of 3 yields near-perfect results, with an RMSE of approximately $1.10 \times 10^{\{-10\}}$, an MAE of $4.93 \times 10^{\{-12\}}$, and an $R^2$ of 1.000, such an outcome is likely indicative of overfitting or data leakage rather than genuine predictive power. To ensure robustness and generalizability, we favor a slightly larger window size. Among the alternatives, a window size of 5 demonstrates excellent performance, with an RMSE of 0.050, an MAE of 0.028, and an $R^2$ of 0.9989. Although there is a minor increase in error compared to the window size of 3, these metrics still reflect a highly accurate and reliable model while reducing the risk of overfitting. As the window size continues to increase (7, 10, 15), a progressive degradation in performance is observed, characterized by higher RMSE and MAE values and a slight decline in $R^2$. Consequently, a window size of 5 is selected as the best compromise between model accuracy and generalization capacity.

The comparison between the cubic trends fitted to $\log(CO_2)$ as a function of $\log(GDP + GDP^2)$ and elasticity reveals notable differences in environmental dynamics is presented in table 2. In the first case, the relationship exhibits a complex cubic Environmental Kuznets Curve (EKC), where CO₂ emissions initially rise with economic growth, stabilize at intermediate levels, and subsequently rise again at very high GDP levels. This non-monotonic behavior highlights potential rebound effects in highly developed economies. In contrast, the relationship between elasticity and $\log(CO_2)$ displays a more conventional inverted U-shaped EKC. Here, CO₂ emissions peak at intermediate elasticity values and decline sharply thereafter, suggesting that greater economic flexibility and adaptation capacities are associated with improved environmental outcomes. These findings imply that elasticity-based analyses may offer clearer insights into the transition towards lower emissions compared to purely income-based models.

Table 2: **The relationship between $CO_2 - GDP$**

| **Global Kutznet $CO_2 - GDP$** | **Global $CO_2 - GDP$ Elasticity** |
|---|---|
| 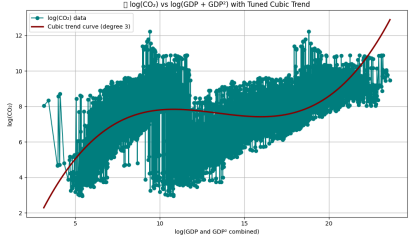 | 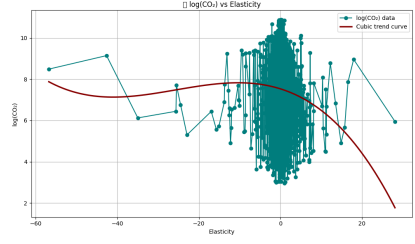 |
| **Kutznet $CO_2 - GDP$ per regions** | **$CO_2 - GDP$ Elasticity per regions** |
| 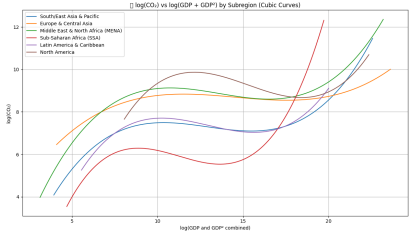 | 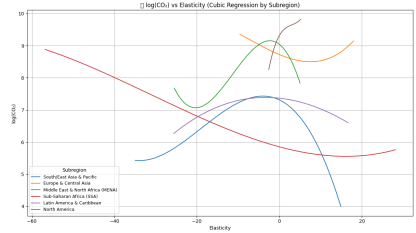 |

Further disaggregation by subregions provides deeper insights. When plotting $\log(CO_2)$ against $\log(GDP + GDP^2)$ by subregion, most areas exhibit a complex cubic relationship. For instance, Sub-Saharan Africa shows a late and steep surge in emissions, while North America and MENA regions initially experience growth followed by a stabilization or mild decline before reacceleration at very high income levels. Conversely, when considering elasticity instead of GDP, the cubic regressions across subregions are more disciplined, with clear peaks and sharper declines, particularly for South/East Asia and Latin America. These patterns suggest that elasticity-based models provide a cleaner and more interpretable structure of the growth–emissions relationship, reinforcing the idea that elasticity captures adaptation dynamics more effectively than raw income levels.

Additional analysis (table 3) exploring the first, second, and third derivatives of elasticity relative to $\log(CO_2)$ further confirms the importance of dynamic economic responses. In each case, the cubic regressions reveal an inverted U-shaped relationship: when the elasticity dynamics are stable (where derivatives close to zero), $CO_2$ emissions are relatively higher, whereas higher rates of change, acceleration, or jerk in elasticity correspond to lower emission levels.

Table 3: Movement of CO2-GDP Elasticity

| **Velocity** | **Acceleration** | **Jerk** |
|---|---|---|
| 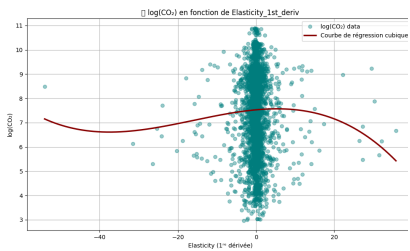 | 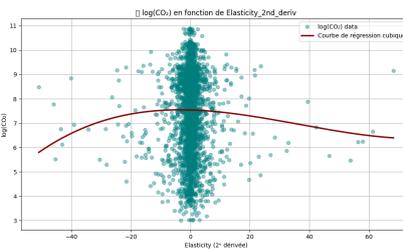 | 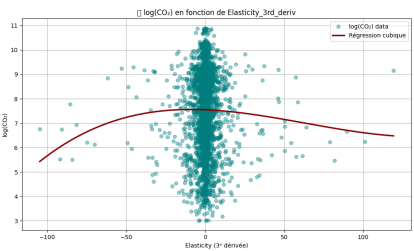 |

These results suggest that it is not only the level of economic flexibility that matters but also the speed and responsiveness of adaptation over time. Economies capable of dynamically adjusting their elasticity at higher rates tend to experience more favorable environmental outcomes. Overall, these findings emphasize the critical role of dynamic, responsive economic structures in achieving sustainable environmental transitions. They look so alike, let us examine their correlation.

The correlation matrix enriched with physical variables highlights several important dynamics. First, while the level of elasticity itself shows negligible correlations with GDP, $CO_2$ emissions, or inflation, its first, second, and third derivatives display strong positive correlations with both GDP and $CO_2$. This suggests that dynamic changes in economic flexibility, rather than static elasticity, are critical in shaping environmental and economic outcomes. Additionally, energy efficiency, measured by $R_{\text{Energ}}$, is negatively correlated with both GDP and $CO_2$, indicating that higher energy efficiency is associated with lower growth rates and reduced emissions. Knowledge and complexity indices (KEI, Inertia, Smoothness) appear to have weaker direct relationships with environmental and economic variables but are moderately associated with economic structures. The internal correlations among elasticity variables reveal the structure of economic dynamic behavior. The elasticity level itself is moderately correlated with its first derivative (0.63) but only with the second (0.32) and third derivatives (0.17). In contrast, the derivatives are strongly interrelated: the first and second derivatives exhibit a correlation of 0.84, and the second and third derivatives reach a correlation of 0.90. These results indicate that while static elasticity remains relatively independent, the dynamic aspects of elasticity—its rate of change, acceleration, and jerk—form a coherent and synchronized system. This suggests that economic adaptations influencing environmental outcomes are driven more by the dynamics of elasticity evolution than by the static levels themselves.

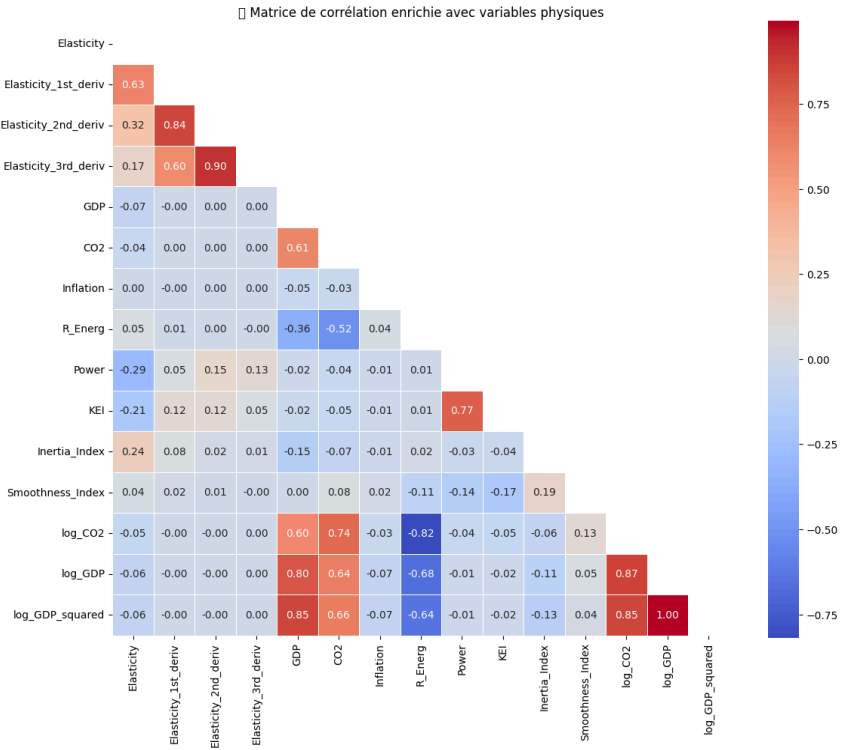

Figure 1: correlation among variables

The above brings more confusion on the question weather first, second and third derivatives of elasticity are not serving the same purpose. Causality analysis understand this new language. The causal network analysis (causality-fig) reveals a tightly organised adjustment mechanism. Economic power exogenously raises elasticity, which cascades through its first, second, and third derivatives, indicating that small shocks propagate rapidly through higher-order dynamics. Elasticity worsens macro-smoothness (p = 0.0319), while kinetic energy (KEI) improves it (p = 0.0001), suggesting that KEI primarily moderates turbulence rather than elasticity itself. Diminished smoothness heightens inflation (p = 0.0008); inflation, in turn, lowers the energy-return ratio, $\log(R_{\text{Energ}})$ (p = 0.0001). Independent of this channel, the jerk of elasticity also erodes energy efficiency (p = 0.0077). Lower energy returns precipitate macroeconomic shocks (p = 0.021), which increase inertia (p = 0.0467); inertia then feeds back negatively on elasticity (p = 0.0012), forming a self-reinforcing loop. The results underscore that dynamic flexibility, not its static level, governs environmental and resilience outcomes, and that policies enhancing knowledge diffusion and energy efficiency can dampen the inflation–shock pathway that locks economies into high-inertia, low-elasticity states.

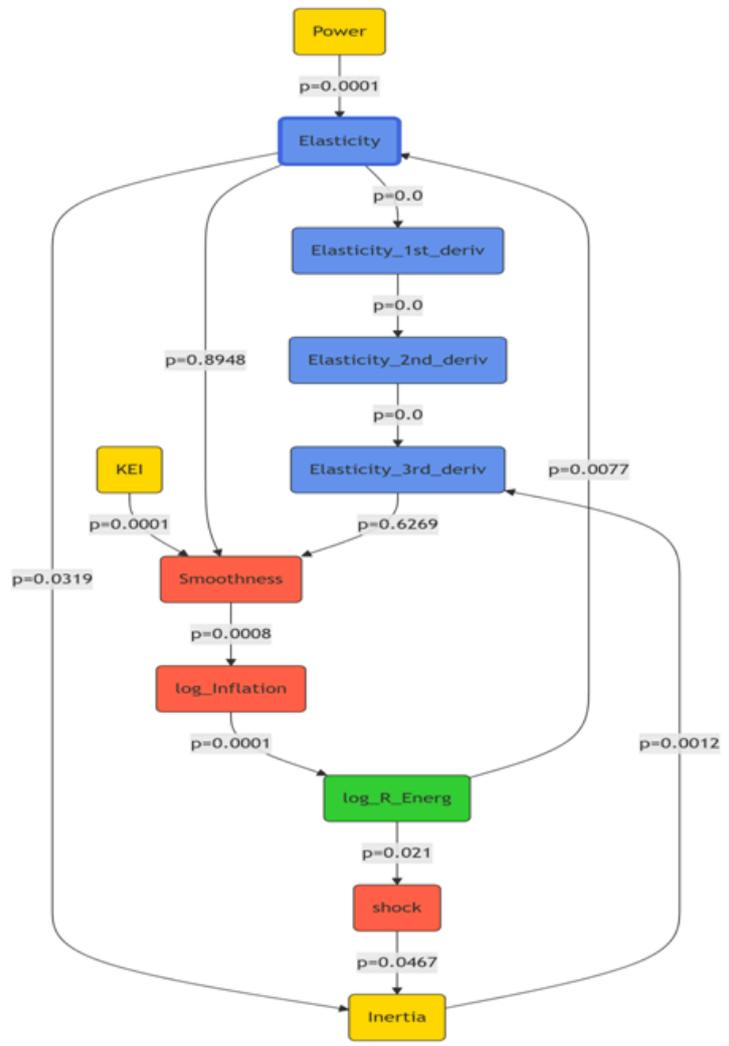

Figure 2: Causality graph among movement indicators

The causal diagram highlights a highly dynamic adjustment process structured around elasticity and its higher-order derivatives.Power initiates the chain by significantly influencing elasticity (p = 0.0001), which sequentially cascades through its first, second, and third derivatives (p = 0.0 at each step), establishing a hierarchical dynamic system where small disturbances at the elasticity level propagate and amplify through velocity, accelerations, and jerks. The third derivative of elasticity (Elasticity_3rd deriv) plays a crucial dynamic role by directly impairing energy return efficiency (p = 0.0077), demonstrating that not the elasticity level but its evolving acceleration patterns critically shape energy outcomes . Additionally, turbulence in the adjustment path, measured through Smoothness, arises partly from elasticity dynamics and is exacerbated or mitigated by knowledge capital (KEI). Lower smoothness feeds into inflationary pressures (p = 0.0008), which further damage energy efficiency (p = 0.0001), creating a dynamic energy degradation loop. The fall in energy returns then triggers macroeconomic shocks (p = 0.021), which reinforce economic inertia (p = 0.0467), ultimately feeding back into the elasticity system itself (p = 0.0012). This closed feedback mechanism underscores that economic adaptability is not merely a function of static flexibility but is profoundly governed by the speed, acceleration, and higher-order dynamics of elasticity adjustments, cascading through inflation, energy efficiency, shocks, and inertia. The above reveals: first, a cascade of causal effects from Elasticity through each derivative. second, causal engines of elasticity which are power and Inertia. Third, control tools notably Smoothness, KEI, and Shock. let us examine implicitly the internal causal structure of dynamics using phase space. Specifically, this analysis inqures how the system's current state (position) causally drives its future change (velocity). This is the language of Hmiltonian theory.

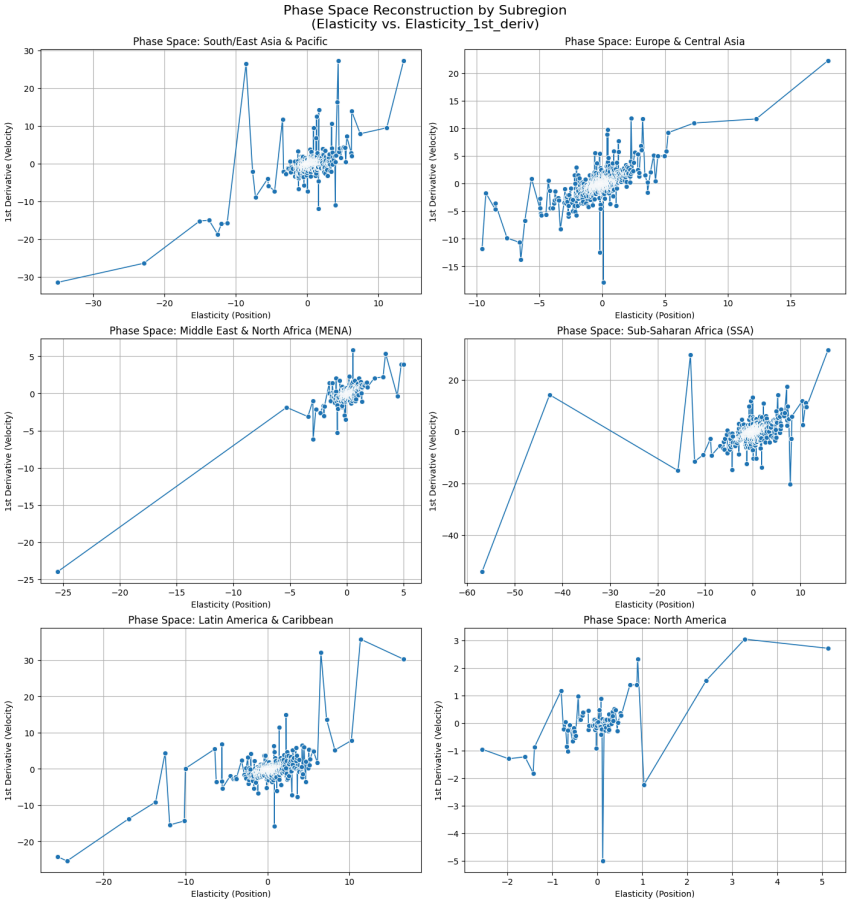

Figure 3: Phase space assessment between elastcity and its Velocity per regions

Hamiltonian theory uses the total energy to predict how things move. Hamiltonian predicts speed and acceleration indirectly, by showing how the energy landscape "pushes" your position and momentum. For instance, the phase space reconstruction of elasticity versus its first derivative reveals profound differences in the dynamic adjustment structures across subregions. Europe and Central Asia, as well as North America, exhibit tightly organized and smooth trajectories, suggesting stable, self-correcting elasticity dynamics that promote macroeconomic resilience. By contrast, South/East Asia, Sub-Saharan Africa, and Latin America display turbulent or chaotic phase spaces, characterized by loops, abrupt velocity changes, and wide swings, indicating heightened vulnerability to external shocks and difficulty in achieving stable adaptations. The Middle East and North Africa (MENA) region shows a distinct pattern of rigidity, with long periods of stability punctuated by abrupt transitions, reflecting fragile adjustment equilibria. These results imply that the resilience of economic systems to shocks is not solely determined by the elasticity level itself, but fundamentally by the dynamic properties of elasticity evolution, as captured by phase space structures.

The phase space reconstruction provides deep insights into the causal structure of elasticity dynamics across subregions. In stable regions such as North America and Europe and Central Asia, elasticity deviations causally induce self-correcting adjustments, consistent with strong negative feedback mechanisms that stabilize the system. In contrast, in regions such as South/East Asia, Sub-Saharan Africa, and Latin America, elasticity deviations causally amplify over time, reflecting positive feedback loops that destabilize the system and promote endogenous shocks. The Middle East and North Africa (MENA) region exhibits threshold-dependent causality, where elasticity remains inert until a critical tipping point triggers abrupt regime shifts. These divergent causal architectures imply that resilience or vulnerability to shocks is fundamentally determined by the dynamic feedback properties inherent in each region's elasticity adjustment process. let us improve this structural appraisal with full Hamiltonian decomposition across regions. But the most interesting is that Hamiltonian doesn't talk about jerk directly, but you can get jerk by taking two time-derivatives of momentum.

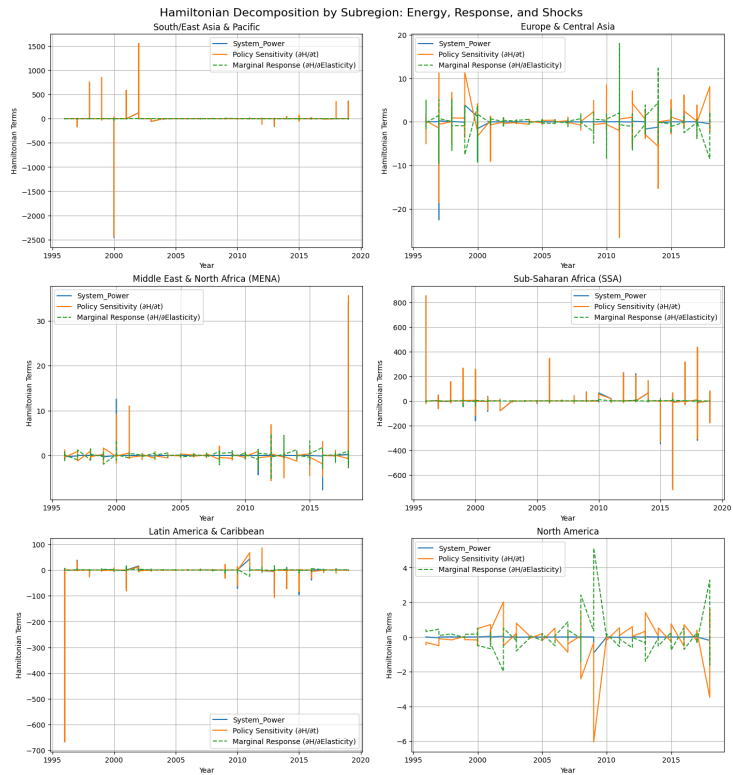

Figure 4: Hamiltonian decomposition of CO2-GDP elasticity across regions

The Hamiltonian decomposition across subregions reveals distinct dynamic profiles of system energy, policy responsiveness, and elasticity-driven marginal responses. South/East Asia and Latin America exhibit highly volatile dynamics, characterized by large surges and collapses in system energy, suggesting that small perturbations in elasticity or policy interventions can trigger systemic crises. In contrast, Europe and Central Asia, and North America maintain stable Hamiltonian structures, where system power and policy sensitivity oscillate within narrow, controlled ranges, reflecting mature stabilization mechanisms. The Middle East and North Africa region demonstrates inert Hamiltonian behavior, with minimal system energy fluctuations, indicating rigidity and limited dynamic adaptability. Sub-Saharan Africa occupies an intermediate position, where moderate shocks and energy swings occur but without full destabilization. These patterns highlight that systemic resilience or fragility is deeply rooted in the underlying Hamiltonian dynamics, with chaotic regions requiring robust stabilization frameworks to prevent energy amplification into crises. an analysis of wavelet scalogram of policy sensitivity by subregion will complete this structural information since it shows how policy response dynamics vary across time and frequency (scale).

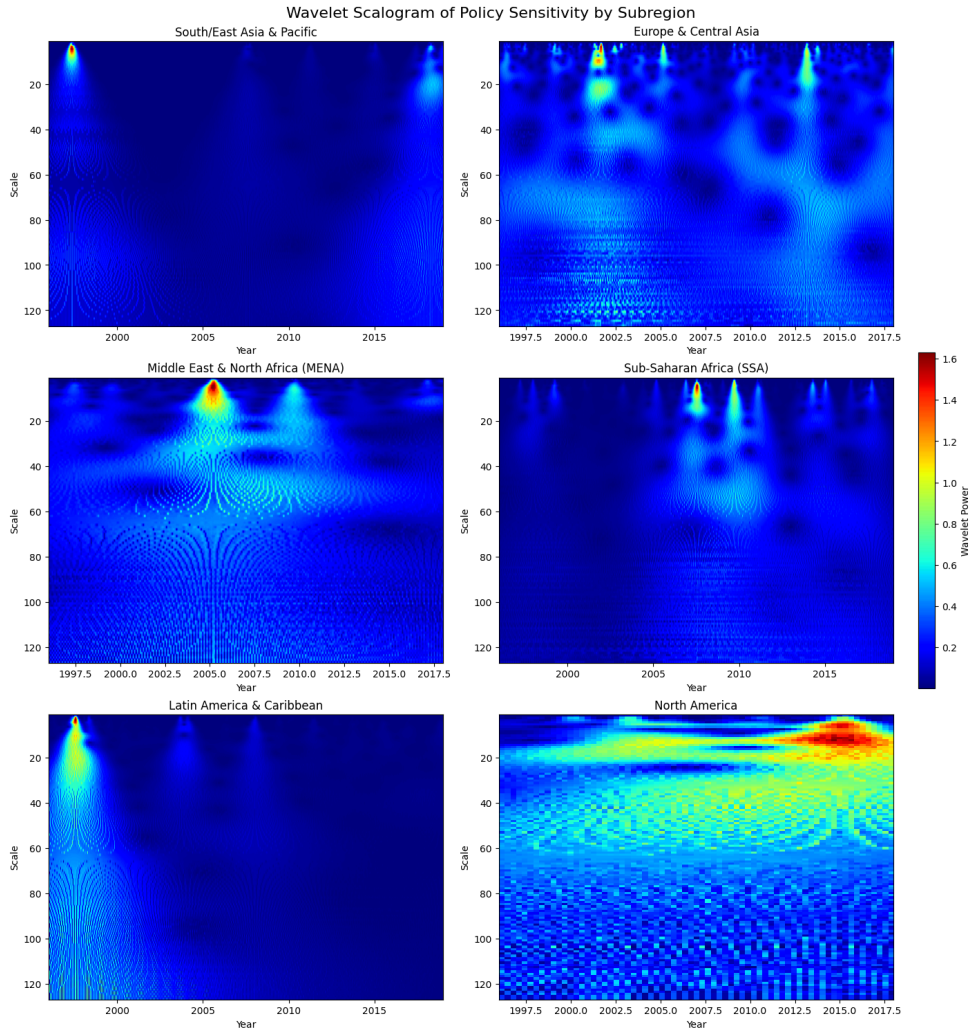

Figure 5: wavelet scalogram diadnosis per region

A wavelet scalogram is a powerful tool for analyzing signals (economic, environmental, or physical data) by decomposing them into time-frequency components. Wavelet scalograms are essential for modern complex systems analysis, particularly in frameworks. By revealing hidden time-frequency structures, they enable precise diagnostics and policy design key for navigating sustainability transitions. They will enable to Expose hidden turning points in decoupling, Optimize policy timing via frequency-specific interventions, or Validate Hamiltonian energy dynamics across scales. It visualizes how different frequencies (or scales) in a signal evolve over time, acting as a "microscope" for dynamic systems. The wavelet scalograms of policy sensitivity reveal markedly different adaptation dynamics across subregions. South/East Asia and Latin America experienced sharp, isolated policy shocks around 2000, concentrated at small scales, suggesting quick but transient adjustment episodes without long-term restructuring. In contrast, Europe and Central Asia, and particularly North America, display sustained wavelet power over time and across scales, indicative of continuous and multiscale policy adaptation capable of managing both short-term disturbances and long-term structural shifts. Middle East and North Africa and Sub-Saharan Africa demonstrate patterns of delayed but significant policy shocks, often with medium-scale persistence, highlighting vulnerabilities to long-term destabilization due to limited continuous adjustment. These findings emphasize that resilience is critically tied to the temporal and frequency structure of policy sensitivity, with continuous multiscale adaptation proving superior to episodic shock absorption.

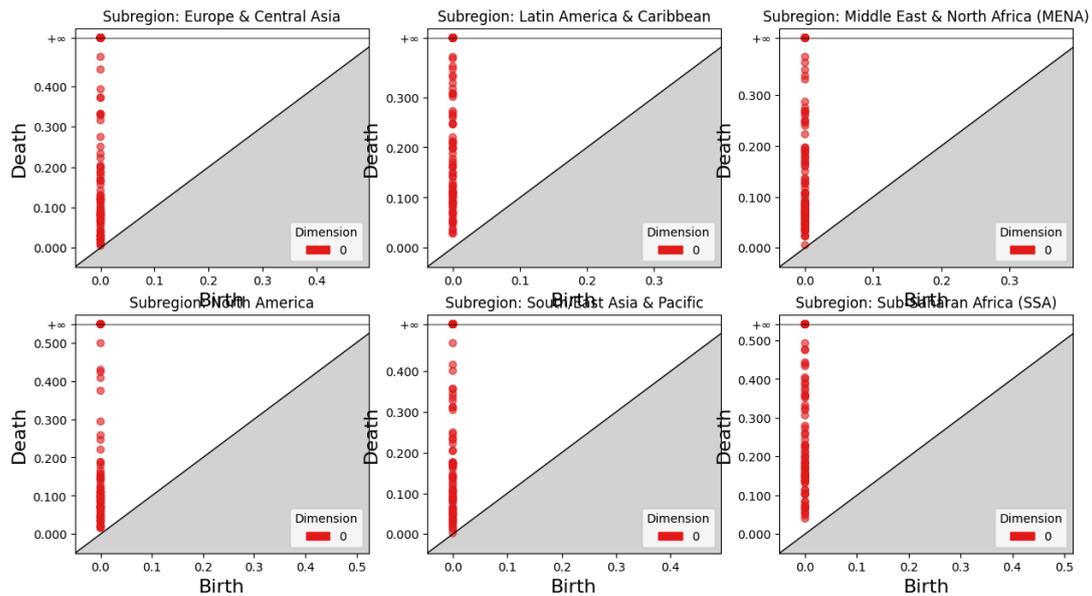

Figure 6: persistence CO2-GDP elasticity per regions

Persistence diagrams are a cornerstone of topological data analysis (TDA), providing a visual and mathematical framework to quantify the evolution of topological features (for example connected components, loops, voids) in data across scales. They are critical for analyzing complex systems like socio-environmental networks, climate dynamics, or economic transitions. This is a 2D plot where each point $(b,d)(b,d)$ represents the birth ($bb$) and death ($dd$) scales of a topological feature in a dataset. Persistence diagrams are essential for decoding the "shape" of complexity in systems like the EKC. By translating data into topological fingerprints, they Expose hidden hierarchies and feedback loops, Provide early warnings of phase transitions, or Enable quantitative comparison of system states. The persistence diagrams for $H\_0$ (connected components) across subregions reveal significant heterogeneity in the underlying topological stability. North America and Sub-Saharan Africa display relatively strong cluster persistence, with points reaching higher death values, indicating the presence of robust, long-lived group structures. In contrast, Europe and Central Asia and Latin America and the Caribbean exhibit highly ephemeral topological features, with clusters emerging and dissolving rapidly, suggesting a fluid and noisy system structure. Middle East and North Africa and South/East Asia and Pacific regions occupy an intermediate position, with some degree of persistent clustering but overall susceptibility to rapid topological shifts. These findings imply that systemic resilience is not only a matter of dynamic adjustment but also rooted in the enduring or transient nature of the system's topological features over time.

It can be very intreersting to redo these analyses per region to bring more time analyses to the table. A million question may also be to know which countries are best fit or can be presented as models in terms of CO2-GDP management? or what is the country with the worse structural model in CO2-GDP mitigation? or which country resist best to CO2-GDP shocks? all these questions and many more can be answered easily using this method.

# Conclusion

This study introduces and applies a novel Hamiltonian Higher-Order Elasticity Dynamics (2HOED) framework to capture the dynamic causal architecture of economic system adaptability. Unlike traditional static analyses, the 2HOED approach models elasticity not only in levels but also through its velocity, acceleration, and jerk, embedding these dynamics within a Hamiltonian structure that accounts for system energy, external shocks, and policy responses. The Hamiltonian decomposition highlighted sharp contrasts across regions, with chaotic explosive dynamics in South/East Asia and Latin America, and stable, finely regulated dynamics in North America and Europe. Phase space reconstructions confirmed these results, showing that stable systems exhibit tight coupling between elasticity, its rate of change, and its acceleration, while chaotic systems display dispersed, nonlinear trajectories consistent with dynamic instability.

Wavelet scalogram analysis of policy sensitivity provided a multiscale validation of the 2HOED framework: regions with continuous multiscale policy adjustments, such as North America and Europe, maintained dynamic energy balance, whereas regions characterized by episodic or delayed policy responses, such as MENA and SSA, exhibited vulnerability to persistent shocks. Persistence diagram analysis further extended the topological interpretation of the 2HOED system, revealing that regions with strong higher-order dynamic regulation also sustain more robust and persistent topological features. Overall, the Hamiltonian Higher-Order Elasticity Dynamics (2HOED) approach proves highly effective in diagnosing the underlying resilience or fragility of economic systems. It shows that economic stability is not solely determined by elasticity levels but by the dynamic interrelations between elasticity trajectories, systemic energy flows, multiscale adaptation, and topological persistence. Future research should build upon this dynamic-topological methodology to explore predictive modeling of crisis onset, system recovery trajectories, and the design of adaptive policy interventions tailored to specific regional dynamic signatures.

# Sample of the study

"Sub-Saharan Africa (SSA)" "Angola", "Benin", "Botswana", "Burkina Faso", "Burundi", "Cameroon", "Cape Verde", "Central African Republic", "Chad", "Comoros", "Congo", "Côte d'Ivoire", "Democratic Republic of the Congo", "Djibouti", "Equatorial Guinea", "Eritrea", "Eswatini", "Ethiopia", "Gabon", "Gambia", "Ghana", "Guinea", "Guinea-Bissau", "Kenya", "Lesotho", "Liberia", "Madagascar", "Malawi", "Mali", "Mauritania", "Mauritius", "Mozambique", "Namibia", "Niger", "Nigeria", "Rwanda", "São Tomé and Príncipe", "Senegal", "Seychelles", "Sierra Leone", "Somalia", "South Africa", "South Sudan", "Sudan", "Tanzania", "Togo", "Uganda", "Zambia", "Zimbabwe"

"Middle East & North Africa (MENA)": [ "Algeria", "Bahrain", "Egypt", "Iran", "Iraq", "Israel", "Jordan", "Kuwait", "Lebanon", "Libya", "Morocco", "Oman", "Palestine", "Qatar", "Saudi Arabia", "Syria", "Tunisia", "Turkey", "United Arab Emirates", "Yemen" ], "Europe & Central Asia": [ "Albania", "Armenia", "Austria", "Azerbaijan", "Belarus", "Belgium", "Bosnia and Herzegovina", "Bulgaria", "Croatia", "Cyprus", "Czech Republic", "Denmark", "Estonia", "Finland", "France", "Georgia", "Germany", "Greece", "Hungary", "Iceland", "Ireland", "Italy", "Kazakhstan", "Kosovo", "Kyrgyzstan", "Latvia", "Lithuania", "Luxembourg", "Malta", "Moldova", "Montenegro", "Netherlands", "North Macedonia", "Norway", "Poland", "Portugal", "Romania", "Russia", "Serbia", "Slovakia", "Slovenia", "Spain", "Sweden", "Switzerland", "Tajikistan", "Turkmenistan", "Ukraine", "United Kingdom", "Uzbekistan" ], "South/East Asia & Pacific": [ "Afghanistan", "Australia", "Bangladesh", "Bhutan", "Brunei", "Cambodia", "China", "Fiji", "India", "Indonesia", "Japan", "Laos", "Malaysia", "Maldives", "Mongolia", "Myanmar", "Nepal", "New Zealand", "North Korea", "Pakistan", "Papua New Guinea", "Philippines", "Samoa", "Singapore", "Solomon Islands", "South Korea", "Sri Lanka", "Taiwan", "Thailand", "Timor-Leste", "Tonga", "Vanuatu", "Vietnam" ], "North America": [ "Canada", "United States", "Mexico" ], "Latin America & Caribbean": [ "Antigua and Barbuda", "Argentina", "Bahamas", "Barbados", "Belize", "Bolivia", "Brazil", "Chile", "Colombia", "Costa Rica", "Cuba", "Dominica", "Dominican Republic", "Ecuador", "El Salvador", "Grenada", "Guatemala", "Guyana", "Haiti", "Honduras", "Jamaica", "Nicaragua", "Panama", "Paraguay", "Peru", "Saint Kitts and Nevis", "Saint Lucia", "Saint Vincent and the Grenadines", "Suriname", "Trinidad and Tobago", "Uruguay", "Venezuela" ]